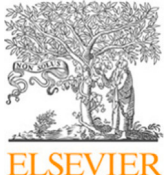
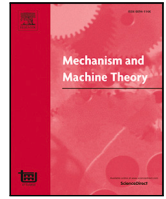

Research paper

# Workspace optimization of 1T2R parallel manipulators with a dimensionally homogeneous constraint-embedded Jacobian

Hassen Nigatu, Doik Kim *

*Center for Intelligent & Interactive Robotics, Korea Institute of Science and Technology (KIST), Seoul, 02792, Republic of Korea*

A B S T R A C T

This paper presents the workspace optimization of one-translational two-rotational (1T2R) parallel manipulators using a dimensionally homogeneous constraint-embedded Jacobian. The mixed degrees of freedom of 1T2R parallel manipulators, which cause dimensional inconsistency, make it difficult to optimize their architectural parameters. To solve this problem, a point-based approach with a shifting property, selection matrix, and constraint-embedded inverse Jacobian is proposed. A simplified formulation is provided, eliminating the complex partial differentiation required in previous approaches. The dimensional homogeneity of the proposed method was analytically proven, and its validity was confirmed by comparing it with the conventional point-based method using a 3-PRS manipulator. Furthermore, the approach was applied to an asymmetric 2-RRS/RRRU manipulator with no parasitic motion. This mechanism has a T-shape combination of limbs with different kinematic parameters, making it challenging to derive a dimensionally homogeneous Jacobian using the conventional method. Finally, optimization was performed, and the results show that the proposed method is more efficient than the conventional approach. The efficiency and simplicity of the proposed method were verified using two distinct parallel manipulators.

## 1. Introduction

One-translational two-rotational (1T2R) type parallel manipulators (PMs) have one translational motion along the *z*-axis and two rotational motions about the x- and y- axes. The remaining axes cannot be used to perform tasks because they are constrained. A typical example of a 1T2R-type PM is a 3-PRS manipulator with a symmetrical limb distribution. However, it is crucial to remember that a small unwanted and uncontrollable displacement, called parasitic motion, occurs along the constrained directions [1]. To eliminate parasitic motion, special limb structures and combinations have been studied [2–4]. These nonparasitic 1T2R PMs have three intersecting limb planes: two planes coincident with each other and the other plane perpendicular to the coincident plane. Hereafter, this type of nonparasitic 1T2R PM is referred to as the T-mechanism to emphasize the shape of the plane intersection.

T-mechanisms are more complex than symmetrical 3-PRS manipulators because of their different limb structures and asymmetrical limb combinations, which makes workspace optimization complex and challenging. The schematics of the two manipulators are shown in Fig. 1.

Increasing the workspace without singularities or at a sufficient distance from the singularity while performing desired tasks is often related to the manipulator's performance optimization. The key step in optimizing PMs is to find an efficient method to determine the workspace. There are several methods for measuring the performance of a manipulator, which are broadly classified as Jacobian-based and non-Jacobian-based approaches [5,6].

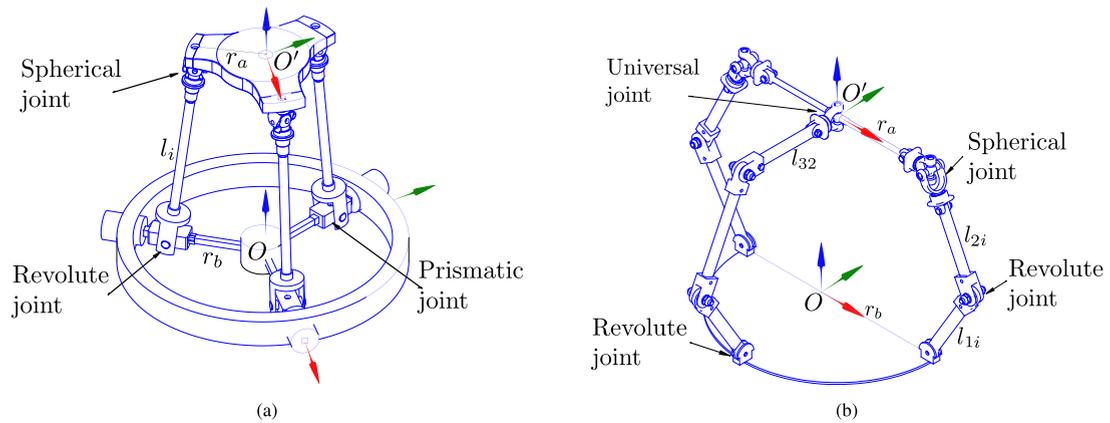

**Fig. 1.** Schematics of example manipulators. (a) 3-PRS manipulator. (b) 2-RRS/RRRU parasitic motion-free manipulator.

Jacobian-based approaches include local and global conditioning indices based on the condition number [7–10], the characteristic length approach [11–13], and the characteristic point-based method [14–16].

Non-Jacobian techniques include angle-based [17], screw theory-based [18,19], and matrix orthogonal degree-based [20] kinematic performance indices.

Although each technique has its own advantages and disadvantages, Jacobian-based approaches appear to be more practical and accessible because most researchers apply a Jacobian to design and analyze PMs in the early stages. However, a Jacobian can only be directly used in optimization if the manipulator's degrees of freedom (DoF) is pure rotation or translation, and its input actuators have a uniform dimension [15,21–26]. The conventional Jacobian matrix of PMs not in this category cannot be used for performance evaluation owing to dimensional inhomogeneity. For mixed DoF manipulators, the terms in the Jacobian are combinations of linear (distance unit) and rotational (dimensionless unit) motion, which leads to inaccurate condition numbers or singular value computations. Moreover, if one switches from radians to degrees, the condition number of the Jacobian gives different values even for an identical manipulator with the same configuration. The actual performance of the manipulator cannot be determined because of the inaccurate and misleading algebraic characteristics of the Jacobian [21]. To deal with this problem, several researchers have proposed methods to homogenize the Jacobian matrix.

Ma and Angeles in 1992, Ranjbaran et al. in 1995, and many others have proposed characteristic length methods [11,27]. However, despite their widespread use, the geometric meaning of such methods is unclear; these are also inconsistent when deriving the characteristic length [16].

In 1992, Gosselin introduced a point-based approach for the planar mechanism [28]. In 2003, Kim and Ryu extended the work of Gosselin by developing an $m \times 9$ Jacobian that mapped nine Cartesian linear velocities of a spatial manipulator into the velocities of $m$-number of actuators, considering that all actuators have a uniform unit [29]. However, suppose three points are selected to represent the motion of a 3- DoF manipulator. Nine elements (three for each point) are obtained and only three are independent while the others are dependent. Because the mapping matrix in [8,29] includes these dependent elements, the computation of the algebraic characteristics may not have physical significance. As a result, the condition number could also cause a misunderstanding of the performance evaluation.

In 2006, Pond et al. [15] proposed a square dimensionally homogeneous Jacobian mapping of only three independent components of selected points on a moving plate to the corresponding actuated joint rate for dexterity analysis. This method is well accepted and can reliably determine the dexterous region of the workspace using the condition number or singular value. However, this approach involves a highly tedious partial differentiation procedure or computational burden if numerical differentiation used.

Liu et al. [14] formulated a point-based square dimensionally homogeneous Jacobian by considering a set of linearly independent axes at the points of a tetrahedron. Although they demonstrated the consistency of their approach using various manipulators, mapping the linear velocity of the selected point to the actuated joint rate is a complex process that in some cases is difficult to understand.

This paper proposes a relatively straightforward method of developing a square dimensionally homogeneous Jacobian for manipulators with mixed DoFs. The proposed method avoids complex, time-consuming procedures and yields an intuitive dimensionally homogeneous Jacobian utilizing a readily available constraint-embedded inverse Jacobian and shifting property. This method has three steps: (1) The manipulator-level inverse rate kinematic relation is obtained based on the analytic reciprocal screw method [30] to explicitly include constraints. The constraint-compatible forward rate kinematic relation is obtained by simply inverting the constraint-embedded inverse Jacobian. (2) The linear velocities of points on the moving plate are obtained using a well-known shifting property [31], and independent elements are filtered by a selection matrix. (3) Finally, by relating the equations in steps 2 and 3, mapping between the independent Cartesian-velocity components and the actuated joint rate, i.e., the square dimensionally homogeneous Jacobian, is achieved.

The method is then applied to optimize the workspace of a 3-PRS [15] and parasitic motion-free 1T2R manipulator [32]. The optimization in this study aims to maximize the workspace by finding the appropriate design variables (e.g., link lengths) using the condition number of the dimensionally homogeneous Jacobian.

The remainder of this paper is organized as follows. Section 2 derives the general dimensionally homogeneous Jacobian. Section 3 describes the determination and optimization of the workspace. Section 4 validates the proposed method using a 3-PRS manipulator by comparing it with the conventional method. In Section 5, a full kinematic relation and dimensionally homogeneous Jacobian of the T-mechanism are derived and optimized. Finally, Section 6 concludes the paper.

## 2. Formulation of dimensionally homogeneous Jacobian

To formulate the dimensionally homogeneous Jacobian, the existing Jacobian for the rate kinematics is modified. First, the inverse rate kinematic relation with constraints is derived as in [30], and then inverted to obtain the constraint-compatible forward rate kinematic relation. Second, three points on the moving plate are selected to represent the entire Cartesian motion of the moving plate. Then, the linear velocities of the points are obtained from the Cartesian motion of the moving plate. The independent components of the point velocities are obtained using a corresponding selection matrix. Finally, by relating the independent linear velocity components of the points on the moving plate and the actuated joint rates, the ($f \times f$) dimensionally homogeneous Jacobin can be obtained, where $f$ denotes the DoF of the manipulator.

*Constraint-embedded rate kinematic relation.* The inverse rate kinematic relation of an $f$-DoF manipulator with embedded constraints is established based on the analytic reciprocal screw relation [30]. The general form is shown as Eq. (1):

$$\begin{bmatrix} \dot{q}_a \\ 0 \end{bmatrix} = \begin{bmatrix} G_a^T \\ G_c^T \end{bmatrix} \dot{x} = G^T \dot{x} \tag{1}$$

where $G_a \in \mathbb{R}^{6 \times f}$ denotes the motion part of the Jacobian and $G_c \in \mathbb{R}^{6 \times (6-f)}$ denotes the constraint part of the Jacobian. $\dot{q}_a \in \mathbb{R}^{f \times 1}$ is the actuated joint rate. $\dot{x} \in \mathbb{R}^{6 \times 1}$ is the Cartesian velocity that must satisfy the structural constraint $G_c^T \dot{x} = 0$.

For any lower DoF manipulator with the above kinematic relation, Eq. (1) can be inverted to obtain the constraint-compatible forward rate relation. Note that $G^T$ can be inverted only if it is nonsingular, i.e., nonsingular in terms of motion and constraint, either independently and/or simultaneously:

$$\dot{x} = J\dot{q} = \begin{bmatrix} J_a & J_c \end{bmatrix} \begin{bmatrix} \dot{q}_a \\ 0 \end{bmatrix} \tag{2}$$

Here, the constraint-compatible forward Jacobian matrix $J \in \mathbb{R}^{6 \times 6}$ is the inverse of $G^T$ and is partitioned for a simplified expression. It can be seen in Eq. (2) that the blocks in the second column are related to the constraint, and thus are multiplied by a zero vector in the joint space. This does not contribute to the generation of actual task motion. Therefore, Eq. (2) can be further simplified as:

$$\dot{x} = J_a \dot{q}_a \tag{3}$$

where $J_a$ has a $(6 \times f)$ dimension.

From Eqs. (1) and (2), the inverse Jacobian $G^T$ and Jacobian $J$ have an inverse relation, and thus:

$$G^T J = \begin{bmatrix} G_a^T \\ G_c^T \end{bmatrix} \begin{bmatrix} J_a & J_c \end{bmatrix} = \begin{bmatrix} G_a^T J_a & G_a^T J_c \\ G_c^T J_a & G_c^T J_c \end{bmatrix} = \begin{bmatrix} I & 0 \\ 0 & I \end{bmatrix} = I \tag{4}$$

where each $I$ and $0$ has a corresponding dimension.

From this relation, the Cartesian velocity $\dot{x}$ obtained from Eq. (2) is constraint-compatible, i.e., it always satisfies the constraints.

*Linear velocity of selected points.* Fig. 2 depicts the moving plate with its Cartesian motion and the selected points. To fully represent the motion of a moving plate in space, it is necessary to identify three non-collinear points embedded on it. These three points can be used to define a plane, which can be described by a vector normal to the plane, denoted as $n$. To describe the pose of the moving plate, one of the three points on the plane ($p_1$) is selected, along with any other point on the plane, denoted as $p$. A plane can then be defined using the vector connecting $p_1$ and $p$, and the normal vector $n$. This plane can be expressed using a plane equation, which can be written as $n^T (p - p_1) = 0$. It is important to note that any point satisfying this plane equation is in the plane, and the plane contains the moving plate, allowing for a full description of its pose.

For most symmetric PMs, the center of the three joints on the moving plate can be taken as these points because they are three and non-collinear. However, for special PMs such as the T-mechanism, the non-collinear points may differ from the center of some of the joints on the moving plate which the details are discussed in Section 5. Nonetheless, Fig. 2 provides an illustration for most cases.

The Cartesian motion $\dot{x}$ passes through the center $O'$ of the moving plate, and is described based on the fixed frame $O$ on the base plate. The center of joints connecting the moving plate with lower links is the most straightforward and logical choice for selecting representative points.

The selected points on the moving plate, denoted by $p_i$ for $i = 1, 2, 3$, can be represented by the position vector $a_i$. However, since the velocity of the points and the moving plate at $O'$ are different, there must be a mapping function to relate them. Using these relations and the shifting property [31], the velocity of the moving plate can be mapped to the linear velocity of a point $p_i$ as follows:

$$\boldsymbol{v}_p = \begin{bmatrix} \boldsymbol{I} & -[\boldsymbol{a}_1]_\times \\ \boldsymbol{I} & -[\boldsymbol{a}_2]_\times \\ \boldsymbol{I} & -[\boldsymbol{a}_3]_\times \end{bmatrix} \dot{\boldsymbol{x}} = \boldsymbol{V}_p \dot{\boldsymbol{x}} \tag{5}$$

where $\boldsymbol{v}_p$ is the linear velocity vector of selected points with 9 dimensions. The matrix $\boldsymbol{V}_p \in \mathbb{R}^{9\times 6}$ is the velocity transition with the $3 \times 3$ identity matrix $\boldsymbol{I}$, and the skew-symmetric matrix $[\boldsymbol{a}_i]_\times$ of the vector $\boldsymbol{a}_i$ where $\times$ is a cross product. The position vector $\boldsymbol{a}_i$ is represented in the inertial coordinate system. Thus, $\boldsymbol{a}_i = \boldsymbol{R}\boldsymbol{a}'_i$, where $\boldsymbol{a}'_i$ is the constant local position vector from $O'$ to a point $p_i$, and $\boldsymbol{R}$ is the orientation matrix of the moving plate.

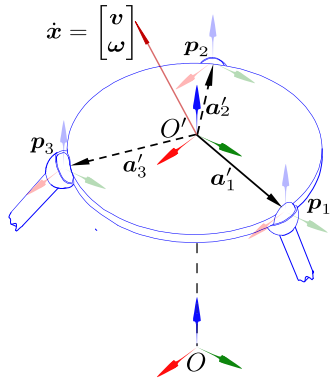

**Fig. 2.** Selected points on the moving plate.

In general, not all components of $\boldsymbol{v}_p$ are independent. Thus, it is important to extract components of $\boldsymbol{v}_p$ that can fully describe the motion of the moving plate with the selection matrix. The selection matrix establishment is dependent on the relation between the linear velocity of points ($\boldsymbol{v}_i$) and moving plate ($\dot{\boldsymbol{x}} = \begin{bmatrix} \boldsymbol{v}^T & \boldsymbol{\omega}^T \end{bmatrix}^T$). This relation is shown in Eq. (5) and expanding it for the $x$, $y$ and $z$ component of the $i$th point yields:

$$\begin{aligned} v_{ix} &= v_x + a_{iz}\omega_y - a_{iy}\omega_z \\ v_{iy} &= v_y - a_{iz}\omega_x + a_{ix}\omega_z \\ v_{iz} &= v_z + a_{iy}\omega_x - a_{ix}\omega_y \end{aligned} \tag{6}$$

From Eq. (6), we can observe that component $v_{ix}$ contains information about $v_x, \omega_x,$ and $\omega_z$. Similarly, $v_{iy}$ comprises $v_y, \omega_x,$ and $\omega_z$, while $v_{iz}$ includes $v_z, \omega_x,$ and $\omega_y$. Any manipulator that has independent motion of $v_x, \omega_x,$ and $\omega_z$ can be uniquely described by $v_{ix}$. Similarly, manipulators that have motion in $v_y, \omega_x,$ and $\omega_z$ or $v_z, \omega_x,$ and $\omega_y$ can be uniquely described with $v_{iy}$ or $v_{iz}$, respectively.

With this relation, we can determine the components that fully describe the motion of 1T2R PMs. Consequently, we can establish the selection matrix to select the independent components of $\boldsymbol{v}_p$ in Eq. (5) as follows:

$$\begin{aligned} \boldsymbol{S}\boldsymbol{v}_p &= \boldsymbol{S}\boldsymbol{V}_p \dot{\boldsymbol{x}} \\ \boldsymbol{v}_{ps} &= \boldsymbol{V}_{ps} \dot{\boldsymbol{x}} \end{aligned} \tag{7}$$

where $\boldsymbol{v}_{ps} = \boldsymbol{S}\boldsymbol{v}_p$ is a vector for the selected independent velocity components, and $\boldsymbol{V}_{ps} = \boldsymbol{S}\boldsymbol{V}_p$ is an $3 \times 6$ matrix to map the selected independent velocity components from the Cartesian velocity of the moving plate.

The selection matrix $\boldsymbol{S} \in \mathbb{R}^{(3\times 9)}$ in Eq. (7) filters the linear velocity components that describe the motion of the moving plate.

In the class of 1T2R parallel manipulators (PMs) involving $z$-axis translation and $x(y)$ rotations, the $z$ component velocity of each point can be chosen to describe the motion of the moving plate, and thus the selection matrix becomes:

$$\boldsymbol{S} = \begin{bmatrix} 0 & 0 & 1 & 0 & 0 & 0 & 0 & 0 & 0 \\ 0 & 0 & 0 & 0 & 0 & 1 & 0 & 0 & 0 \\ 0 & 0 & 0 & 0 & 0 & 0 & 0 & 0 & 1 \end{bmatrix} \tag{8}$$

While this paper primarily focuses on the $T_z R_x R_y$ type of 1T2R PMs, it is important to note that the complete enumeration of 1T2R PMs comprises nine varieties, including $T_x R_x R_y$, $T_x R_x R_z$, $T_x R_y R_z$, $T_y R_x R_y$, $T_y R_x R_z$, $T_y R_y R_z$, $T_z R_x R_y$, and $T_z R_x R_z$.

From these nine varieties, only $T_x R_y R_z$, $T_y R_x R_z$, and $T_z R_x R_y$ PMs can be described using $v_{ix}$, $v_{iy}$, and $v_{iz}$, respectively. For other six manipulators, a combination of them must be used to fully describe the motion of the moving plate (See Appendix B for the detail formulation of selection matrix for this group of PMs).

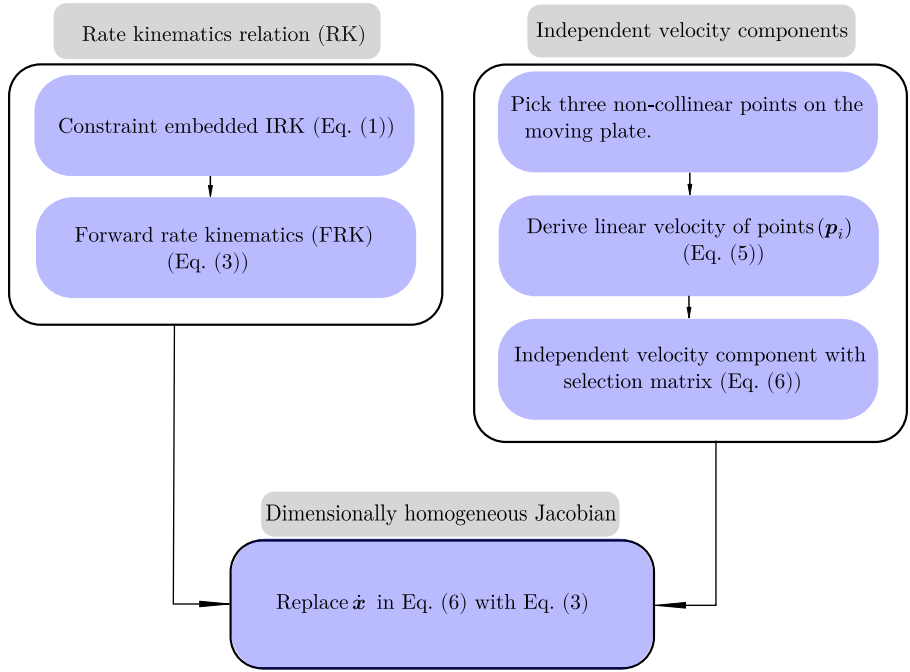

**Fig. 3.** Procedure for dimensionally homogeneous Jacobian.

*Dimensionally homogeneous Jacobian.* Now, we relate the linear velocity of the points on the moving plate to the joint rate by substituting Eq. (3) into Eq. (7):

$$\boldsymbol{v}_{ps} = \boldsymbol{V}_{ps}\dot{\boldsymbol{x}} = \boldsymbol{V}_{ps}\boldsymbol{J}_a\dot{\boldsymbol{q}}_a = \boldsymbol{J}_{dh}\dot{\boldsymbol{q}}_a \tag{9}$$

The product of $\boldsymbol{V}_{ps} \in \mathbb{R}^{(f \times 6)}$ and $\boldsymbol{J}_a \in \mathbb{R}^{(6 \times f)}$ is an $f \times f$ matrix, i.e., the dimensionally homogeneous Jacobian, $\boldsymbol{J}_{dh} = \boldsymbol{V}_{ps}\boldsymbol{J}_a$. It can be seen that Eq. (9) is only dependent on the constraint-compatible forward Jacobian. Thus, no new parameters or procedures are needed in the intermediate process. To summarize the procedures described above, please refer to the flow of procedure shown in Fig. 3.

Before using $\boldsymbol{J}_{dh}$ in Eq. (9), its dimensional homogeneity is discussed. From Eq. (5), $\boldsymbol{I}$ is dimensionless, whereas $\boldsymbol{a}_i$ has a unit of length. If the Cartesian velocity $\dot{\boldsymbol{x}}$ is multiplied, then the result gives a unit of linear velocity. Consequently, if we obtain the correct forward Jacobian, we can obtain the dimensionally homogeneous mapping of $\boldsymbol{J}_{dh}$ in Eq. (9).

The unit of the inverse Jacobian $\boldsymbol{G}^T$ is known because it is derived analytically in this study. To determine the unit of the forward Jacobian $\boldsymbol{J}$, the inverse Jacobian is inverted analytically using the approach in [33]. First, matrix $\boldsymbol{G}^T$ is partitioned into $2 \times 2$ submatrices as:

$$\boldsymbol{G}^T = \begin{bmatrix} \boldsymbol{G}_{av}^T & \boldsymbol{G}_{aw}^T \\ \boldsymbol{G}_{cv}^T & \boldsymbol{G}_{cw}^T \end{bmatrix} \tag{10}$$

Because $\boldsymbol{G}_{av}$ is nonsingular, the inverse of Eq. (10) is obtained as follows:

$$(\boldsymbol{G})^{-T} = \boldsymbol{J} = \begin{bmatrix} \boldsymbol{G}_{av}^{-T} + \boldsymbol{G}_{av}^{-T}\boldsymbol{G}_{aw}^T(\boldsymbol{G}_{cw}^T - \boldsymbol{G}_{cv}^T\boldsymbol{G}_{av}^{-T}\boldsymbol{G}_{aw}^T)^{-1}\boldsymbol{G}_{cv}^T\boldsymbol{G}_{av}^{-T} & \times \\ -(\boldsymbol{G}_{cw}^T - \boldsymbol{G}_{cv}^T\boldsymbol{G}_{av}^{-T}\boldsymbol{G}_{aw})^{-1}\boldsymbol{G}_{cv}^T\boldsymbol{G}_{av}^{-T} & \times \end{bmatrix} = \begin{bmatrix} \boldsymbol{J}_{a1} & \times \\ \boldsymbol{J}_{a2} & \times \end{bmatrix} \tag{11}$$

where $(\cdot)^{-T} = ((\cdot)^T)^{-1}$. As the second column in Eq. (11) is not of interest here, we simply ignore it and focus on $\boldsymbol{J}_a$.

The unit relation is dependent on the inverse Jacobian; hence, we must specify a manipulator to analyze a unit relation. For example, assume that a manipulator has rotational actuators, and the inverse Jacobian in Eq. (10) has the following unit relation: $\boldsymbol{G}_{av}$ and $\boldsymbol{G}_{cv}$ have an inverse of length (1/m), and $\boldsymbol{G}_{aw}$ and $\boldsymbol{G}_{cw}$ are dimensionless. From Eq. (11), we can deduce that $\boldsymbol{J}_{a1}$ has a unit of length (m), while $\boldsymbol{J}_{a2}$ has no unit (i.e., dimensionless). The Cartesian velocity $\dot{\boldsymbol{x}}$ obtained from the above unit relationship provides the correct dimensionally homogeneous Jacobian with Eq. (9). This relationship will be evaluated in Sections 4 and 5 using specific manipulators.

## 3. Workspace determination and optimization

By using the condition number or determinant of the dimensionally homogeneous Jacobian $\boldsymbol{J}_{dh}$ in Eq. (9), the geometric parameters of a manipulator can be optimized. In other words, by maximizing the condition number with a threshold, the dexterous workspace volume can be maximized; or by setting the determinant det($J_{dh}$) = 0, the reachable workspace can be maximized. This section describes the determination and optimization of the workspace with the dimensionally homogeneous Jacobian.

*3.1. Workspace determination*

The workspace of 1T2R PMs is typically defined by three-dimensional Cartesian coordinates, which is the pose achievable with the origin of the moving plate. The workspace volume is a function of the geometric parameters and structural constraints imposed on the moving plate.

Most workspace determination algorithms for PMs are of a discretization type [34–36], in which a grid of the positions and orientations of the points is defined via the sampling step. Smaller sampling steps lead to improved accuracy in determining the workspace boundary. An existing numerical discretization-based workspace determination algorithm in [35–37] was used to validate the proposed dimensionally homogeneous Jacobian. First, the pose parameters are discretized with an evenly arranged grid in the polar coordinate form. Each point in the grid is checked to determine whether it falls inside the workspace.

The condition number of the dimensionally homogeneous Jacobian derived in Section 2 is used to achieve the desired performance by keeping the manipulator away from the singularity. Algorithm 1 briefly describes how the workspace is determined using the $J_{dh}$ in Eq. (9). In the algorithm, $\Delta z$ is the small height difference between the lowest height $z_0$ and the highest height $z_f$. $\Delta \alpha$ and $\Delta \epsilon$ are the orientation changes in the polar coordinates for the angle change and direction change, respectively. Thus, $\alpha$ is changed to determine the workspace boundary, and $\epsilon$ is searched for one revolution, $2\pi$.

$n$ and $m$ are the number of grids for the height and meridians, respectively. The height is divided by $n$ to obtain $\Delta z = (z_f - z_0)/n$, and the search angle direction is divided by $m$ to obtain $\Delta \epsilon = 2\pi/m$. The initial angle difference $\Delta \alpha_0$ was obtained independently from $\Delta \epsilon$. However, for convenience, $\Delta \alpha_0$ is given as the same value $\Delta \alpha_0 = \Delta \epsilon$. The angle difference $\Delta \alpha$ was changed by reducing it to half its value to obtain a more exact workspace boundary. The resolution of the workspace boundary is given by th$_{boundary}$. $\Delta \phi$ and $\Delta \psi$ are the small increments of the roll and pitch angles, respectively. $k$ is the condition number of $J_{dh}$, while $k_{max}$ is its maximum value set as a threshold.

The workspace boundary detection can be summarized by the following step-by-step process.

- The first step is to divide the heave motion into $n$ steps, which are the vertical slices of the total volume. A slice represents an area with reachable orientation angles at a particular height.
- A radial search is applied along the direction ($\epsilon$) of the independent orientation angle with a smaller step size ($\Delta \alpha$). The search continues in the polar coordinate until the prescribed condition number limit ($k_{max}$) is reached. When the condition number exceeds the limit, the step size is divided in half, and the orientation angle decreases. Then, the search continues in the same direction until the increment becomes smaller than the tolerance th$_{boundary}$.
- Once the tolerance and condition number threshold are reached along that particular direction, the orientation angles are recorded as the boundary of the workspace.
- The increment $\Delta \epsilon$ is then applied to determine the new radial direction ($\epsilon$), and the search re-starts from zero angles ($\psi = \phi = 0$). This increment is completed after a full rotation, i.e., $2\pi$ is reached. At this point, the complete workspace of a slice is determined.
- Finally, the heave or height is updated by a vertical increment $\Delta z$ until it reaches the maximum height.

*3.2. Workspace optimization*

In this study, the maximum workspace volume with the maximum possible orientation angles and heave movements are achieved by changing the values of the geometric parameters. In the optimization, one can determine the reachable or dexterous workspace [36]. Here, we opted to maximize the dexterous workspace with the appropriate threshold of the condition number.

Therefore, it is necessary to quantify the workspace volume. As shown in the previous section, the workspace was calculated with three inputs: two orientation angles ($\psi, \theta$) and the height $z$ of the 1T2R PMs. Moreover, the condition number of the dimensionally homogeneous Jacobian with given structural parameters, such as the radii of the base and moving plate and link lengths, was used. All of these parameters can be the design values $\rho$ in the optimization. The generalized equation of the cost function is expressed as:

$$\max_{\rho^*} \quad V(\psi, \theta, z, \rho) \qquad (12)$$
$$\text{s.t.} \quad \text{cond}(J_{dh}) < k_{max}$$

where $k_{max}$ is the maximum condition number as a constraint to remain in the dexterous workspace.

The workspace volume is obtained from Algorithm 1 as follows:

$$V = \sum_{i=0}^{n} \sum_{j=0}^{m} \left( \frac{\Delta \epsilon}{2} \sqrt{\psi_{i,j}^2 + \theta_{i,j}^2} \; \Delta z_i \right) \qquad (13)$$

Finally, the *pattern search* algorithm of Matlab® was used to isolate the optimal value from a set of data points within the given search domain to obtain the maximum dexterous volume. *Pattern search* is a nongradient optimization method that uses a search pattern around the existing points. This method was adopted because it ensures global convergence.

**Algorithm 1:** Workspace boundary determination

```
input : z_0, z_f, n, m, Δα_0, Δε ← 2π/m, Δz ← (z_f - z_0)/n, th_boundary, k_max
output: z, ψ, θ
for i = 0 : n do
    ε ← 0                                                       ▷ Reset values
    z = z_0 + iΔz
    for j = 0 : m do
        ψ ← 0, θ ← 0                                            ▷ Reset values
        Δα ← Δα_0                                               ▷ Reset search angle difference
        while Δα > th_boundary do
            k ← 0
            Δθ ← Δα sin(ε)                                      ▷ Calculate angle difference
            Δψ ← Δα cos(ε)
            ψ ← ψ - Δψ                                          ▷ Modify angles for initialization
            θ ← θ - Δθ
            while k < k_max do
                θ ← θ + Δθ                                      ▷ Calculate angles for Jacobian
                ψ ← ψ + Δψ
                Compute J_dh                                    ▷ Jacobian and condition number.
                k ← cond(J_dh)
            end
            ψ ← ψ - Δψ                                          ▷ Make angles inside workspace
            θ ← θ - Δθ
            Δα ← Δα/2                                           ▷ Reduce search angle difference
        end
        ψ(i,j) ← ψ                                              ▷ Save workspace boundary angles
        θ(i,j) ← θ
        ε ← ε + Δε                                              ▷ Change search direction
    end
    z(i,j) ← z                                                  ▷ Save height
    Plot z, ψ, θ
end
```

## 4. Optimization of 3-PRS manipulator for validation

### 4.1. Workspace determination

The 3-PRS manipulator shown in Fig. 1(a) was originally presented in [1]. It is a symmetrical 1T2R mechanism with parasitic motion in its workspace. The position and velocity-level inverse relation of the manipulator used in this section was derived in [38]. The dimensionally homogeneous Jacobian of this mechanism was obtained based on the approach given in Section 2. Note that the choice of points on the moving plate for this manipulator follows the general rule described in Section 2. Then, the workspace was determined based on the Algorithm 1 setting $n = m = 150$, $k_{max} = 6$, $z_0 = 0.001$, and $z_f = 1$; and the initial link lengths $l_1 = l_2 = l_3 = 1$. The workspace shape and volume obtained in this study are exactly the same as those in [15,36]. Thus, the proposed dimensionally homogeneous Jacobian can be used for optimization, and is expected to yield results that are consistent with previous results. The initial dexterous workspace obtained using the proposed method is shown in Fig. 4.

### 4.2. Optimal dexterous workspace

Based on the report in [39], the relevant design variables for this mechanism are the radius of the moving plate ($r_a$), link lengths ($l_i$), and inclination angle from the base plate that determines the line of action of the prismatic joint ($\gamma$). Thus, the design variable vector as in Eq. (12) is $\rho = [r_a \; l_i \; \gamma]^T$. The upper and lower limits of the parameters are $0.1 \leq r_a \leq 1$, $0.1 \leq l \leq 1$, and $0 \leq \gamma \leq 90$ deg. Optimization was performed along with the maximum permissible condition number $k_{max} = 6$ as a constraint. The resulting optimal dexterous workspace is shown in Fig. 5 with the optimal parameters $r_a = 0.620$, $l_i = 1$, and $\gamma = 0$. The optimization of this manipulator using the dimensionally homogeneous Jacobian proposed in this study was compared with the method used in [15], as shown in Tables 1 and 2, respectively. The tables list the initial design variables($\rho_0$), initial volume ($V_0$), optimal design variable($\rho_{opt}$), optimal volume($V_{opt}$), iteration number(Iter.) and computation time(sec).

The results show that the proposed method requires less computational time because it is based on an analytical inverse Jacobian. The longest computation time required by the method in [15] was 22,200 s, while the proposed method required 14,372 s, a reduction of 35% on an Intel(R) 1.80 GHz Core(TM) i7-8550U computer.

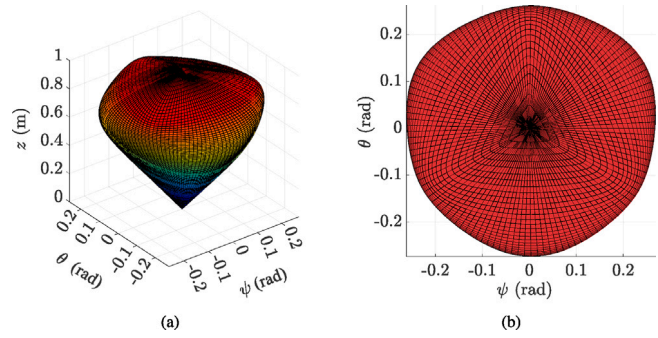

**Fig. 4.** The dexterous workspace of a 3-PRS manipulator with $k_{max} = 6$. (a) Workspace 3D view. (b) Workspace 2D view (top).

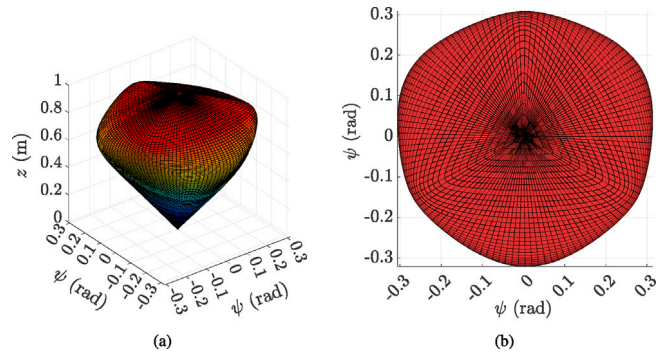

**Fig. 5.** Optimal dexterous workspace of a 3-PRS manipulator. (a) Workspace 3D view. (b) Workspace 2D view (top).

**Table 1**
Optimized workspace volume: proposed method.

|  | $\rho_0$ | $V_0$ | $\rho_{opt}$ | $V_{opt}$ | Iter. | sec |
|---|---|---|---|---|---|---|
|  | 0.4, 0.4, 0.0 | 0.0132 | 0.640,1.0,0.0 | 0.09978 | 120 | 10,973 |
| $r_a, l_i, \gamma$ | 0.2, 0.2, 0.2 | 0.0109 | 0.639,1.0,0.0 | 0.09975 | 139 | 12,507 |
|  | 0.1, 0.1, 0.1 | 0.0120 | 0.620,1.0,0.0 | 0.09977 | 161 | 14,472 |

**Table 2**
Optimized workspace volume: method in [15].

|  | $\rho_0$ | $V_0$ | $\rho_{opt}$ | $V_{opt}$ | Iter. | sec |
|---|---|---|---|---|---|---|
|  | 0.4, 0.4, 0.0 | 0.0132 | 0.640,1.0,0.0 | 0.09976 | 186 | 16,950 |
| $r_a, l_i, \gamma$ | 0.2, 0.2, 0.2 | 0.0109 | 0.639,1.0,0.0 | 0.09977 | 211 | 19,320 |
|  | 0.1, 0.1, 0.1 | 0.0120 | 0.620,1.0,0.0 | 0.09977 | 249 | 22,200 |

Note that the work in [36] also determined the $3 \times 3$ square Jacobian matrix for optimizing the 3-PRS mechanism, but it was not dimensionally homogeneous. Nonetheless, it did provide a stable condition number for this particular manipulator by setting its linear unit in meters. This is not a common but a special case. For other cases, such as 3-RRS manipulators, the method in [36] cannot be used, although the methods in [15,39] are needed to obtain the dimensionally homogeneous Jacobian. The method proposed in this study can be applied to all cases with a dimensionally homogeneous Jacobian.

## 5. Optimization of T-mechanism of 2-RRS/RRRU

The T-mechanism of the 2-RRS/RRRU structure is a class of 1T2R manipulators whose basic structure comes from the 3-RRS mechanism by removing the undesirable parasitic motion [1,2,4,32]. A class of T-mechanisms can cover three-legged parasitic motion-free PMs with different leg configurations.

The T-mechanism has an asymmetric architecture and is more complex than the 3-PRS mechanism described in Section 4. Moreover, it has never been analyzed in detail. Thus, Section 5.1 discusses the kinematic features in detail.

## 5.1. Dimensionally homogeneous Jacobian

Fig. 1(b) shows the schematics of the T-mechanism. The angles between each leg from the positive *x*-axis in the clockwise direction are given by $\xi = [0, -90°, -180°]$. The pose of the moving plate is denoted by the position $p$ and orientation $R$. Limbs 1 and 3 have an identical RRS joint arrangement, but limb 2, which is orthogonal to limbs 1 and 3, has an RRRU joint arrangement. Limbs 1 and 3 have two links denoted by $l_{11}(l_{13})$ and $l_{21}(l_{23})$, respectively. Limb 2 has three links: $(l_{12}, l_{22}, $ and $l_{32})$. All revolute joints in each limb are parallel and represented by the direction vector $s_{ji\|}$, where $j$ and $i$ are the joint and limb numbers, respectively. Limb 1 is coincident with the *x*-axis of the fixed frame. The constants $r_a$ and $r_b$ represent the radii of the moving and base plates, respectively. They are the distance between the joints connecting the limbs and the corresponding plate. The vector $a_i$ represents the position of a joint on the moving plate with respect to the fixed frame, and is also represented as $a_i = Ra'_i$ with the local constant position vector $a'_i$. Similarly, $b_i$ is a joint on the base plate.

The T-mechanism has no parasitic motion; hence, the workspace is represented only by the independent variables of translation along the *z*-axis and rotation about the x- and y- axes. The output rotation was decoupled because limb 2 generated the *x*-axis rotation ($\psi$), while the coordination of limbs 1 and 3 generated the *y*-axis orientation ($\theta$) independently. As previously discussed, obtaining the inverse Jacobian is a fundamental factor in formulating the dimensionally homogeneous Jacobian. The inverse rate kinematics of the T-mechanism, whose detailed derivation is shown in Appendix A, was rewritten with a conventional representation of the Cartesian velocity $\dot{x} = [v^T; w^T]^T$, which has exchanged the order of velocities in Eq. (A.29) as follows:

$$\begin{bmatrix} \dot{q}_a \\ 0 \end{bmatrix} = \begin{bmatrix} \frac{l_{21}^T}{l_{21} \cdot (s_{11\|} \times l_{11})} & \frac{(l_{21} \times a_1)^T}{l_{21} \cdot (s_{11\|} \times l_{11})} \\ \frac{l_{22}^T}{s_{12\|} \cdot (l_{12} \times l_{22})} & k\frac{(s_{42\|} \times s_{52\|})^T}{s_{12\|} \cdot (l_{12} \times l_{22})} \\ \frac{l_{23}^T}{l_{23} \cdot (s_{13\|} \times l_{13})} & \frac{(l_{23} \times a_3)^T}{l_{23} \cdot (s_{13\|} \times l_{13})} \\ s_{11\|}^T & (s_{11\|} \times a_1)^T \\ s_{12\|}^T & 0^T \\ s_{13\|}^T & (s_{13\|} \times a_3)^T \end{bmatrix} \begin{bmatrix} v \\ w \end{bmatrix} = \begin{bmatrix} G_a^T \\ G_c^T \end{bmatrix} \dot{x} \quad (14)$$

By inverting Eq. (14), the forward rate kinematic relation of Eq. (3) can be obtained, which will be used for incorporating the constraints.

Since the 3-RRS/RRRU (T-mechanism) has 3-DoF, three noncolinear points are sufficient to describe the motion of the moving plate. Unfortunately, the three points connecting the limbs and moving plate are colinear in this case; therefore, we need at least one noncolinear point. Two points of limbs 1 and 3 were selected, i.e., the center of the spherical joints. The remaining point relating limb 2 and the moving plate was selected somewhere on link $l_{32}$. As shown in Fig. 6, one candidate point can be $p_2$, temporarily coincident with the third revolute joint of limb 2. It is worth noting that in our approach, we used the constraint-embedded Jacobian relation to derive the dimensionally homogeneous Jacobian. Hence, the points must be the distal-link of the limbs. Moreover, the point at the third revolute joint of limb 2 is always parallel to the moving plate due to the nature of universal joint. Therefore, it is important to consider that if the second limb's last joint differed, a distinct choice would be required, and further analysis might be necessary.

Accordingly, the linear velocities of the selected points are given by:

$$v_p = \begin{bmatrix} v_1 \\ v_2 \\ v_3 \end{bmatrix} = \begin{bmatrix} I_3 & -[a_1]_\times \\ I_3 & -[l_{32}]_\times \\ I_3 & -[a_3]_\times \end{bmatrix} \begin{bmatrix} v \\ \omega \end{bmatrix} = V_p \dot{x} \quad (15)$$

where $l_{32} = RT_y(-l_{32})$ and $l_{32}$ is the length of the selected points from $O'$, and $v_i$ is the linear velocity of the selected point $i$.

Now, we need to select the independent components from the linear velocities of the selected points. For the T-mechanism presented here, the *z*-directional velocity of each selected point was independent and sufficient to describe the motion of the moving plate. Thus, we multiply Eq. (8) with Eq. (15), to obtain the following equation that relates the independent linear velocity of the points to $\dot{x}$:

$$v_{ps} = SV_p\dot{x} = V_{ps}\dot{x} = \begin{bmatrix} 0 & 0 & 1 & a_{1y} & -a_{1x} & 0 \\ 0 & 0 & 1 & l_{32y} & -l_{32x} & 0 \\ 0 & 0 & 1 & a_{3y} & -a_{3x} & 0 \end{bmatrix} \begin{bmatrix} v \\ \omega \end{bmatrix} \quad (16)$$

Finally, by substituting $\dot{x}$ in Eq. (16) with the inverse relation of Eq. (14), the $(3 \times 3)$ dimensionally homogeneous Jacobian, $J_{dh}$ for the T-mechanism is obtained as Eq. (9):

$$v_{ps} = V_{ps}J_a\dot{q}_a = J_{dh}\dot{q}_a \quad (17)$$

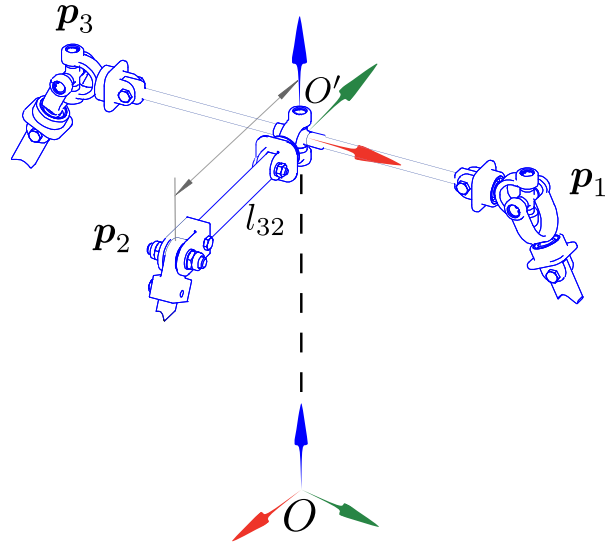

**Fig. 6.** Selected representative points on the moving plate.

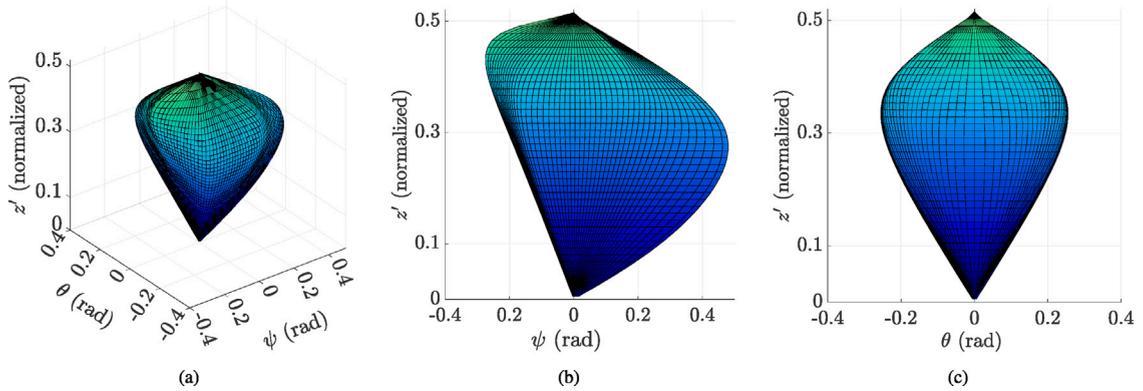

**Fig. 7.** Initial workspace of the T-mechanism with $k_{max} = 2$. (a) Workspace 3D view. (b) 2D view ($\psi - z$ plane). (c) 2D view ($\theta - z$ plane).

### 5.2. Optimization of the T-mechanism

The dexterous workspace of the manipulator was optimized using the square dimensionally homogeneous matrix $J_{dh}$ given in Eq. (17) and the procedure in Algorithm 1. The initial parameters used to obtain the workspace shown in Fig. 7 are $r_b = r_a = l_{1i} = l_{2i} = l_{32} = 45$ mm. The condition number used to limit the workspace is $k_{max} = 2$. The number of grids is $n = 150$ and $m = 150$.

As shown in Fig. 7, the workspace with the initial parameters has a small orientation range and must be optimized. It should be noted that the height in Fig. 7 was normalized with the optimized link lengths to compare the results before and after optimization.

Before optimization, the design parameters influencing the workspace were rearranged. Because limbs 1 and 3 are being coupled and symmetrical, the same parameters were assigned to limb 1. The parameters for limbs 1 and 3 are the link lengths $l_{11} = l_{13}$, $l_{21} = l_{23}$; those for limb 2 are the link lengths $l_{12}$, $l_{22}$, $l_{32}$. For the plates, the radii of the moving plate $r_a$ and base plate $r_b$ are also design parameters. We added one more parameter $r_{b2}$ as the radius of the base plate for limb 2 because it is decoupled from other limbs, and it may be helpful to increase the workspace with different limb positions. This is optional but was added in the optimization for generality. Finally, the radius of the base plate $r_b$ is the reference length, and all other lengths are represented as relative ratios to remove the size effect. This leaves only seven parameters for optimization as follows:

$$l_{11} = l_{13} = \rho_1 r_b, \quad l_{21} = l_{23} = \rho_2 r_b, \quad l_{12} = \rho_3 r_b, \quad l_{22} = \rho_4 r_b, \quad l_{32} = \rho_5 r_b, \quad r_{b2} = \rho_6 r_b, \quad r_a = \rho_7 r_b \tag{18}$$

In the optimization, the ratio vector $\rho = [\rho_1 \ \rho_2 \ \rho_3 \ \rho_4 \ \rho_5 \ \rho_6 \ \rho_7]^T$ was used. For the initial workspace shown in Fig. 7, the coefficients are one, which implies that all parameters have the same value as $r_b$.

The maximum height of the manipulator is the full extension of limbs 1 or 3, $z_{max} = l_{1i} + l_{2i}$. Thus, the actual height $z$ was normalized to remove the size effect as $z' = z/z_{max}$. The normalized height range $z'$ is always $0 \leq z' \leq 1$. Finally, the cost function

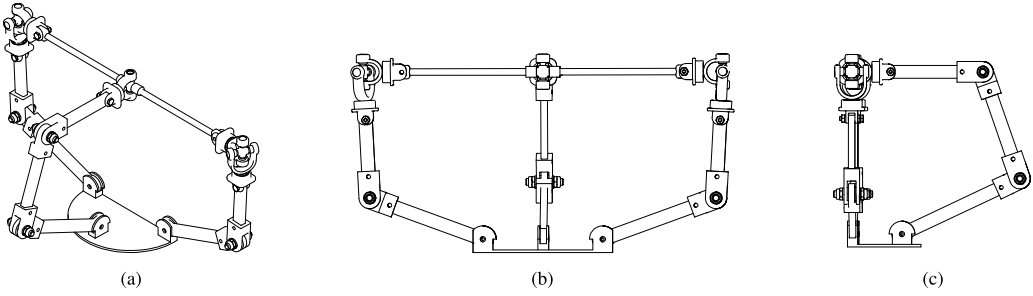

**Fig. 8.** Optimal T-mechanism. (a) 3-D view. (b) x-z plane view. (c) y-z plane view.

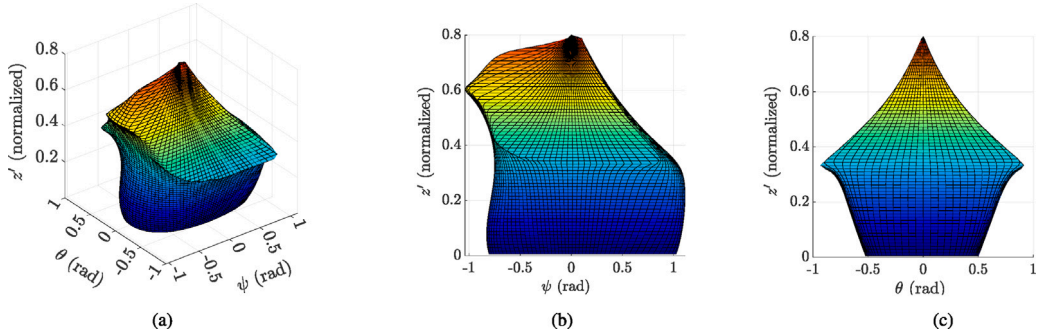

**Fig. 9.** Optimal workspace of T-mechanism with $k_{\max} = 2$. (a) Workspace 3D view. (b) 2D view ($\psi - z$ plane). (c) 2D view ($\theta - z$ plane).

**Table 3**
Dexterous workspace volume optimization: Method I.

|  | $\rho_0$ | $V_0$ | $\rho_{opt}$ | $V_{opt}$ | Iter. | sec |
|---|---|---|---|---|---|---|
| $\rho_1, \rho_2, \rho_3, \rho_4, \rho_5, \rho_6, \rho_7$ | 1, 1, 1, 1, 1, 1, 1 | 0.051 | 2.19, 1.72, 2.24, 1.94, 1.50, 0.84, 3.00 | 0.48182 | 510 | 45,960 |
|  | 2, 2, 2, 2, 2, 2, 2 | 0.01 | 2.19, 1.72, 2.24, 1.94, 1.50, 0.84, 3.08 | 0.48091 | 673 | 60,539 |
|  | 0.5, 0.5, 0.5, 0.5, 0.5, 0.5, 0.5 | 0.0014 | 2.19, 1.72, 2.24, 1.94, 1.50, 0.84, 3.00 | 0.48391 | 660 | 59,292 |

Eq. (12) was changed to a normalized formulation as:

$$\max_{\rho^*} \quad V(\psi, \theta, z', \rho)$$
$$\text{s.t.} \quad \text{cond}(\boldsymbol{J}_{dh}) < k_{\max} \quad (19)$$

The maximum condition number is $k_{\max} = 2$. The coefficient ratio is bounded in the range $0.01 \leq \rho \leq 10$.

*Method I: Full-set optimization.* First, all the seven coefficients in $\rho$ were simultaneously optimized with Eq. (13) and Eq. (19). The following optimal values are obtained.

$$\boldsymbol{\rho}_{opt} = \begin{bmatrix} 2.0092 & 1.7207 & 2.2441 & 1.9355 & 1.5000 & 0.8398 & 3.000 \end{bmatrix} \quad (20)$$

From the optimized link ratio $\boldsymbol{\rho}_{opt}$, the optimized link lengths (mm) can be obtained by multiplying the reference length $r_b = 45.0$ mm as follows:

$$\boldsymbol{l}_{opt} = \boldsymbol{\rho}_{opt} r_b = \begin{bmatrix} l_{11}(l_{13}) & l_{21}(l_{23}) & l_{12} & l_{22} & l_{32} & r_{b2} & r_a \end{bmatrix}$$
$$= \begin{bmatrix} 90.4 & 77.4 & 101.0 & 87.1 & 67.5 & 37.8 & 135.0 \end{bmatrix} \quad (21)$$

The home configuration of the manipulator with optimized parameters is shown in Fig. 8. The optimization result shows that $l_{11} = l_{13} > l_{21} = l_{23}$, $l_{12} > l_{22} > l_{32}$, $r_{b2} < r_b$, and $r_a > r_b$.

The resulting optimal workspace is shown in Fig. 9. Fig. 9(a) shows a 3D view of the workspace. Fig. 9(b) and (c) depict the $\psi - z$ and $\theta - z$ plane views, respectively. As shown in the figure, the workspace is much larger than the original one in Fig. 7. Table 3 shows the results of three optimization trials with different initial parameters.

For the purpose of comparison, the T-mechanism was also optimized using the method in [15]. The results are listed in Table 4. Similar to Section 4, the proposed method is more efficient in terms of the optimization time and number of iterations, as shown in Tables 3 and 4.

**Table 4**
Dexterous workspace volume optimization: Based on [15].

| | $\rho_0$ | $V_0$ | $\rho_{opt}$ | $V_{opt}$ | Iter. | sec |
|---|---|---|---|---|---|---|
| $\rho_1, \rho_2, \rho_3, \rho_4, \rho_5, \rho_6, \rho_7$ | 1, 1, 1, 1, 1, 1, 1 | 0.051 | 2.18, 1.72, 2.25, 1.93, 1.50, 0.84, 3.00 | 0.47610 | 602 | 54,859 |
| | 2, 2, 2, 2, 2, 2, 2 | 0.01 | 2.18, 1.71, 2.23, 1.94, 1.50, 0.84, 3.08 | 0.48012 | 831 | 75,727 |
| | 0.5, 0.5, 0.5, 0.5, 0.5, 0.5, 0.5 | 0.0014 | 2.19, 1.73, 2.25, 1.94, 1.50, 0.84, 3.00 | 0.48391 | 818 | 74,543 |

**Table 5**
Dexterous workspace volume optimization: Method II.

| | $\rho_0$ | $V_0$ | $\rho_{opt}$ | $V_{opt}$ | Iter. | sec |
|---|---|---|---|---|---|---|
| $\rho_1, \rho_2, \rho_7$ | 1, 1, 1 | 0.051 | 2.01, 1.72, 3.00 | 0.209412 | 138 | 12,405 |
| | 2, 2, 2 | 0.010 | 2.01, 1.72, 3.00 | 0.209422 | 133 | 12,002 |
| | 0.5, 0.5, 0.5 | 0.0014 | 2.01, 1.72, 3.01 | 0.209401 | 136 | 12,207 |
| $\rho_3, \rho_4, \rho_5, \rho_6$ | 1, 1, 1, 1 | 0.21 | 2.24, 1.94, 1.50, 0.84 | 0.483911 | 68 | 6202 |
| | 2, 2, 2, 2 | 0.21 | 2.24, 1.94, 1.50, 0.84 | 0.483911 | 71 | 6360 |
| | 0.5, 0.5, 0.5, 0.5 | 0.21 | 2.24, 1.94, 1.50, 0.84 | 0.483907 | 67 | 6210 |
| $\rho_1, \rho_2, \rho_3, \rho_4, \rho_5, \rho_6, \rho_7$ | Optimal values[a] | 0.48 | 2.19, 1.72, 2.24, 1.94, 1.50, 0.84, 3.0 | 0.483900 | 20 | 1800 |

[a] All optimal values obtained from phase one and phase two are used to perform complete optimization.

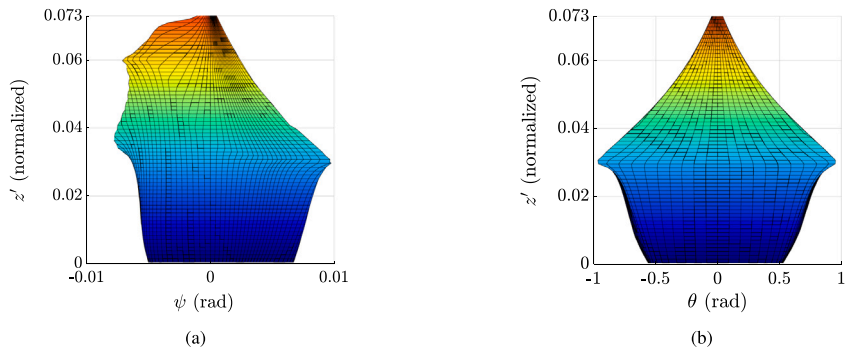

**Fig. 10.** Intermediate workspace with $k_{max} = 2$ after optimizing $\rho_1, \rho_2$ and $\rho_7$. (a) 2D view ($\psi - z$ plane). (b) 2D view ($\theta - z$ plane).

*Method II: Decoupled-set optimization.* In Section 5.1, it was shown that the T-mechanism has a decoupled motion, i.e., limbs 1 and 3 for the $y$-axis rotation and limb 2 for the $x$-axis rotation, independently. Using this decoupled property can further improve the optimization performance. To do so, the parameters were separately grouped as follows: parameters associated with limbs 1 and 3 ($\rho_1, \rho_2, \rho_7$), and parameters of limb 2 ($\rho_3, \rho_4, \rho_5, \rho_6$). These two groups of parameters were optimized separately. Then, these separately optimized parameters were again optimized simultaneously to obtain the final result.

The number of parameters in the first group of parameters was much smaller than the total number of parameters; therefore, their optimization was more efficient. When the first group of parameters was optimized, some of the other parameters were also optimized because they shared the same ratio. Thus, the second group of parameters were optimized more efficiently, and the final overall optimization started from almost optimized parameters. Consequently, the overall optimization quickly converged to the solution. Table 5 lists the optimization results with the same initial conditions in Table 3.

The intermediate workspace in Fig. 10, obtained before optimizing second group of parameters, shows that after the optimization of limbs 1 and 3, the $x$-axis rotation ($\psi$) of the manipulator was reduced compared with the initial rotation shown in Fig. 7(b). The valid heave range slightly increased but was less than the optimal height shown in Fig. 9. The $y$-axis rotation ($\theta$) is similar to the final optimized rotation shown in Fig. 9, i.e., the $y$-axis rotation was optimized, as expected.

Although the final results are identical regardless of whether the parameters were optimized simultaneously or separately, the optimization time significantly decreased, as shown in Tables 3 and 5. For example, the first initial parameter case of the full-set optimization in Table 3 took 510 iterations in 45,960 s, but the decoupled-set optimization in Table 5 took only 226 iterations in 20,407 s. The T-mechanism has a complex kinematic structure, However its decoupled motion makes the optimization more efficient.

*Optimized workspace with different conditions.* To observe how the workspace size and shape varied with the increase in the condition number, Fig. 11 was plotted with $k_{max} = 2, 6,$ and $\infty$. Figs. 11(a) and 11(b) are obtained with $k = 2$ and $k = 6$, respectively. For convenience, $k_{max} = \infty$ is approximated with $\det(\mathbf{J}_{dh}) = 0$ as shown in Fig. 11(c). It can be seen that the workspace increased as the condition number increased. The shape became more symmetrical compared with that of the lower condition number.

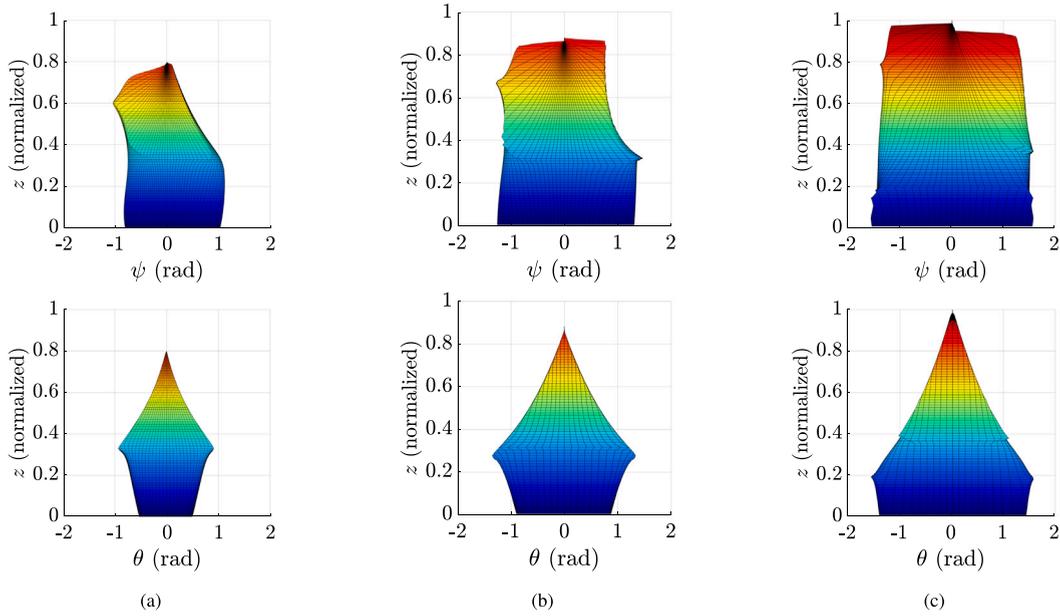

**Fig. 11.** Optimal dexterous workspace 2D view. (a) $k_{max} = 2$. (b) $k_{max} = 6$. (c) $k_{max} = \infty$.

## 6. Conclusion

This study derived a constraint-embedded dimensionally homogeneous Jacobian for workspace optimization. The dimensionally homogeneous Jacobian was obtained from an existing inverse Jacobian without separate derivations by applying a point-based approach and selecting the independent velocities. Its dimensional homogeneity was also analytically proven. The proposed method reduced the complexity of the derivation process and the computational burden. The method was applied to 3-PRS and 2-RRS/RRRU manipulators. The 3-PRS manipulator was used for comparison, and the optimization results were equivalent to previous results. The T-mechanism was used to validate the workspace optimization performance of a nonsymmetric and complex manipulator. First, seven kinematic parameters (three for the RRS, four for the RRRU) were selected and optimized simultaneously to maximize the workspace. Compared with the previous method, the computational burden was reduced by 20%. Further, by using the decoupled structure of the T-mechanism, the optimization process was divided into the following: (1) the optimization of RRS, (2) the optimization of RRRU, and 3) and optionally, the optimization of all the parameters simultaneously. The results were consistent with those of simultaneous optimization. However, the total computation time was significantly reduced to an average of 224 iterations (20,262 s), which is only 36.5% of the time of simultaneous optimization and 30% of that of the previous method. The proposed method can also be applied to optimize other lower DoF PMs with different constraints and motions.

## Declaration of competing interest

The authors declare that they have no known competing financial interests or personal relationships that could have appeared to influence the work reported in this paper.

## Data availability

Data will be made available on request.

## Acknowledgment

This work was supported by the Korea Institute of Science and Technology (KIST) Institutional Program (2E31581).

## Appendix A. Derivation of inverse rate kinematics of the t-mechanism

As discussed in Section 5, the manipulator has two RRS and one RRRU limbs. Consequently, the Jacobians of these limbs are separately derived. The inverse Jacobian of the RRS limb is first derived, and then that of the RRRU limb is formulated. Finally, the manipulator-level inverse Jacobian is established from the limb-level inverse Jacobian. A limb-level Jacobian is a Jacobian from the origin of the base plate $O$ to the end of the limb that connects the limb and moving plate, i.e., a spherical joint for limbs 1 and 3, and a universal joint for limb 2. A manipulator-level Jacobian combines these three limb-level Jacobians into one by multiplying the corresponding transformation, which transforms the end of the limb into the origin of the moving plate $O'$.

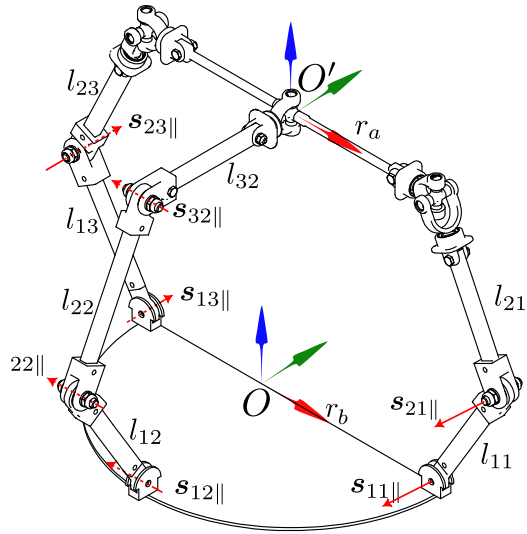

**Fig. A.1.** Schematics of the T-mechanism.

In the following sections, all the derivation processes are described with the screw representation, i.e., $\$ = [w^T; v^T]^T$, for convenience.

*A.1. Rate kinematics of limb RRS*

Based on the schematics in Fig. A.1, the Jacobian of the two identical RRS limbs, limbs 1 and 3, is obtained as:

$$J_s = \begin{bmatrix} s_{1s\parallel} & s_{2s\parallel} & s_{3s\parallel} & s_{4s\parallel} & s_{5s\parallel} \\ s_{1s\parallel} \times r_{1s} & s_{2s\parallel} \times l_{2s} & 0 & 0 & 0 \end{bmatrix} \tag{A.1}$$

where $s_{is\parallel}$ is the direction vector for the $i$th joint of the limb $s = 1, 3$; and $r_{is}$ is the distance from joint $i$ to the end of the limb. For example, $r_{1s}$ is the distance from joint 1 to the end of the limb, i.e., $r_{1s} = l_{1s} + l_{2s}$. The position vector $l_{2s}$ can be rewritten as $r_{2s}$ with this definition. Note that the direction of joint 1, $s_{1s\parallel}$, is parallel to that of joint 2, $s_{2s\parallel}$, i.e., $s_{1s\parallel} \parallel s_{2s\parallel}$. In addition, the position vector can be considered as $r_{3s} = r_{4s} = r_{5s} = 0$ from Eq. (A.1). The limb restriction screw is obtained from the following constraint equation established from the 5 zero-pitch, 5$\$_0$, screw system of Eq. (A.1) as:

$$\begin{bmatrix} s_{5s\parallel}^T & 0^T \\ s_{4s\parallel}^T & 0^T \\ s_{3s\parallel}^T & 0^T \\ s_{2s\parallel}^T & (s_{2s\parallel} \times l_{2s})^T \\ s_{1s\parallel}^T & (s_{1s\parallel} \times r_{1s})^T \end{bmatrix} \begin{bmatrix} s_{r\perp} \\ s_{r\parallel} \end{bmatrix} = 0 \tag{A.2}$$

The reciprocal screw satisfying Eq. (A.2) is obtained by the analytical methods in [30,40]. The direction vector for this limb is analytically obtained for the 5$\$_0$ system as follows:

$$\begin{aligned}
s_{r\parallel} = &-[s_{5s\parallel} \cdot (s_{4s\parallel} \times s_{3s\parallel})]\big((p_{52} \times s_{2s\parallel}) \times (p_{51} \times s_{1s\parallel})\big) \\
&+[s_{5s\parallel} \cdot (s_{3s\parallel} \times s_{1s\parallel})]\big((p_{54} \times s_{4s\parallel}) \times (p_{52} \times s_{2s\parallel})\big) \\
&-[s_{5s\parallel} \cdot (s_{4s\parallel} \times s_{1s\parallel})]\big((p_{53} \times s_{3s\parallel}) \times (p_{52} \times s_{2s\parallel})\big) \\
&+[s_{5s\parallel} \cdot (s_{3s\parallel} \times s_{2s\parallel})]\big((p_{54} \times s_{4s\parallel}) \times (p_{51} \times s_{1s\parallel})\big) \\
&-[s_{5s\parallel} \cdot (s_{4s\parallel} \times s_{5s\parallel})]\big((p_{54} \times s_{4s\parallel}) \times (p_{53} \times s_{3s\parallel})\big) \\
&+[s_{5s\parallel} \cdot (s_{4s\parallel} \times s_{2s\parallel})]\big((p_{53} \times s_{3s\parallel}) \times (p_{51} \times s_{1s\parallel})\big)
\end{aligned} \tag{A.3}$$

where $p_{5i} = p_5 - p_i$ which implies a vector between the 5th and $i$th joint screws; thus, it is equal to $p_{5i} = r_{5s} - r_{is}$. From the relation in Eq. (A.1), $p_{54} = p_{53} = 0$ because the position vector $r_{3s} = r_{4s} = r_{5s} = 0$.

Eq. (A.3) is simplified as:

$$
\begin{aligned}
s_{r\|} &= -[s_{5s\|} \cdot (s_{4s\|} \times s_{3s\|})]\left((p_{52} \times s_{2s\|}) \times (p_{51} \times s_{1s\|})\right) \\
&\approx (p_{52} \times s_{2s\|}) \times (p_{51} \times s_{1s\|}) \quad \because \text{dropping scalar value} \\
&= (l_{2s} \times s_{2s\|}) \times ((l_{1s} + l_{2s}) \times s_{1s\|}) \quad \because p_{52} = -l_{2s} \text{ and } p_{51} = -r_{1s} = -(l_{1s} + l_{2s}) \\
&= (l_{2s} \times s_{2s\|}) \times (l_{1s} \times s_{1s\|}) \quad \because s_{1s\|} \parallel s_{2s\|} \\
&= (l_{2s} \cdot (s_{2s\|} \times s_{1s\|}))l_{1s} - (l_{2s} \cdot (s_{2s\|} \times l_{1s}))s_{1s\|} \quad \because \text{vector quadruple product} \\
&\approx s_{1s\|} \quad \because s_{1s\|} \parallel s_{2s\|} \text{ and dropping scalar value}
\end{aligned}
\tag{A.4}
$$

By substituting the direction vector $s_{r\|}$ obtained in Eq. (A.4) into Eq. (A.2), the moment vector $s_{r\perp}$ can be solved, and turns out to be zero, $s_{r\perp} = \mathbf{0}$. Then, the limb restriction screw becomes as in Eq. (A.5):

$$
\$_{cs} = \begin{bmatrix} s_{1s\|}^T & \mathbf{0}^T \end{bmatrix}^T \tag{A.5}
$$

By embedding the restriction screw, the extended Jacobian becomes:

$$
J_{es} = \begin{bmatrix} s_{1s\|} & s_{2s\|} & s_{3s\|} & s_{4s\|} & s_{5s\|} & \mathbf{0} \\ s_{1s\|} \times r_{1s} & s_{2s\|} \times l_{2s} & \mathbf{0} & \mathbf{0} & \mathbf{0} & s_{1s\|} \end{bmatrix} \tag{A.6}
$$

where the last column is the instantaneous constraint that cannot exhibit motion in the direction. All the others are related to the joint motion.

The inverse relation of the active joint, i.e., the first joint, is obtained by removing the first column of Eq. (A.6) to have five screw systems. Then, the reciprocal screw of the first joint can be obtained similarly as in the above procedure for the restriction screw (see [30,40,41] for a detailed description). The final result is:

$$
\$_{as} = \begin{bmatrix} l_{2s}^T & \mathbf{0}^T \end{bmatrix}^T \tag{A.7}
$$

Similarly, the reciprocal screws of the other passive joints can be obtained.

Now we have the limb-level inverse Jacobian of the RRS branch. By changing the reference point, the relation can be transformed into the manipulator-level inverse Jacobian. The process to transform the reference point from the center of the spherical joint to the origin of the moving plate is:

$$
\mathbf{M} = \begin{bmatrix} \mathbf{I} & \mathbf{0} \\ -[a_s]_\times & \mathbf{I} \end{bmatrix} \tag{A.8}
$$

where $a_s$ is the vector from the spherical joint to the origin of the moving plate and $[\cdot]_\times$ is the skew-symmetric matrix of vector $(\cdot)$.

Consequently, we obtained the transformed (manipulator-level) wrenches as:

$$
w_{as} = \begin{bmatrix} (l_{2s} \times a_s)^T & l_{2s}^T \end{bmatrix}^T \tag{A.9}
$$

$$
w_{cs} = \begin{bmatrix} (s_{1s\|} \times a_s)^T & s_{1s\|}^T \end{bmatrix}^T \tag{A.10}
$$

Note that the wrenches are reciprocal coordinates with different orders.

### A.2. Rate kinematics of limb RRRU

The limb-level Jacobian of the second limb is obtained as:

$$
J_2 = \begin{bmatrix} s_{12\|} & s_{22\|} & s_{32\|} & s_{42\|} & s_{52\|} \\ s_{12\|} \times r_{12} & s_{22\|} \times r_{22} & s_{32\|} \times r_{32} & \mathbf{0} & \mathbf{0} \end{bmatrix} \tag{A.11}
$$

where $r_{12} = l_{12} + l_{22} + l_{32}$, $r_{22} = l_{22} + l_{32}$ and $r_{32} = l_{32}$.

Following a similar procedure as that for limb RRS, the limb restriction screw is obtained from Eq. (A.11) as shown below.

$$
\$_{c2} = \begin{bmatrix} s_{12\|}^T & \mathbf{0}^T \end{bmatrix}^T \tag{A.12}
$$

The extended Jacobian by adding the restriction screw is:

$$
J_2 = \begin{bmatrix} s_{12\|} & s_{22\|} & s_{32\|} & s_{42\|} & s_{52\|} & \mathbf{0} \\ s_{12\|} \times r_{12} & s_{22\|} \times r_{22} & s_{32\|} \times r_{32} & \mathbf{0} & \mathbf{0} & s_{12\|} \end{bmatrix} \tag{A.13}
$$

The reciprocal screw for the active joint is obtained by removing the first column of the extended Jacobian Eq. (A.13), which is the $4\$_0 - 1\$_\infty$ screw system, as follows:

$$
\begin{bmatrix} s_{52\|}^T & \mathbf{0}^T \\ s_{42\|}^T & \mathbf{0}^T \\ s_{32\|}^T & (s_{32\|} \times l_{32})^T \\ s_{22\|}^T & (s_{22\|} \times (l_{22} + l_{32}))^T \\ \mathbf{0}^T & s_{12\|}^T \end{bmatrix} \begin{bmatrix} s_{a\perp} \\ s_{a\|} \end{bmatrix} = \mathbf{0} \tag{A.14}
$$

The reciprocal screw satisfying Eq. (A.14) is obtained by the analytic methods in [30,40]. The direction vector for this limb is obtained analytically for the $4\$_0 - 1\$_\infty$ system as follows:

$$
\begin{aligned}
s_{a\parallel} = &-[s_{52\parallel} \cdot (s_{32\parallel} \times s_{22\parallel})]\Big((p_{54} \times s_{42\parallel}) \times s_{12\parallel}\Big) \\
&+[s_{52\parallel} \cdot (s_{42\parallel} \times s_{22\parallel})]\Big((p_{53} \times s_{32\parallel}) \times s_{12\parallel}\Big) \\
&-[s_{52\parallel} \cdot (s_{42\parallel} \times s_{32\parallel})]\Big((p_{52} \times s_{22\parallel}) \times s_{12\parallel}\Big)
\end{aligned}
\tag{A.15}
$$

where $p_{54} = 0$ is the position vector $r_{42} = r_{52} = 0$. The other vectors are $p_{53} = -l_{32}$ and $p_{52} = -(l_{22} + l_{32})$.

Using the relations of the vector triple product, parallel direction vectors of revolute joints $s_{12\parallel} \parallel s_{22\parallel} \parallel s_{32\parallel}$, and the orthogonality of the link vector and direction vector of the revolute joints $p_{5i} \perp s_{j2\parallel}$ for $i, j = 1, 2, 3$, Eq. (A.15) is simplified as $s_{a\parallel} = l_{22}$.

The moment vector in this case is more complex than the limb RRS, which must satisfy Eq. (A.14). From the first and second rows in Eq. (A.14), we obtain the moment vector perpendicular to both $s_{52\parallel}$ and $s_{42\parallel}$ of the universal joint as:

$$ s_{a\perp} = k(s_{42\parallel} \times s_{52\parallel}) $$

but the constant scalar $k$ is not specified. The constant $k$ can be obtained from the third row as follows:

$$ s_{32\parallel} \cdot s_{a\perp} = k s_{32\parallel} \cdot (s_{42\parallel} \times s_{52\parallel}) = -(s_{32\parallel} \times l_{32}) \cdot l_{22} \tag{A.16} $$

$$ k = \frac{s_{32\parallel} \cdot (l_{22} \times l_{32})}{s_{32\parallel} \cdot (s_{42\parallel} \times s_{52\parallel})} \tag{A.17} $$

Note that the fourth row is equivalent to the third row in terms of the relation of the scalar triple product, the orthogonality between the revolute joint direction vector and link vector, and the parallel condition of the revolute joint direction vector. The fifth row always yields zero with the orthogonality between the revolute joint direction vector and link vector.

The universal joint can have several types of installations with different joint directions. However, from Eq. (A.17), the universal joint direction, $s_{42\parallel}$ and $s_{52\parallel}$ need not be parallel with the joint direction $s_{32\parallel}$ to avoid a zero denominator. Consequently, the universal joint must be installed with the joints heading toward the y- and z-axes, as shown in Fig. A.1.

No transformation is required for this limb because its terminal is directly attached to the origin of the moving plate. The wrenches for the active joint and constraints are:

$$ w_{a2} = \begin{bmatrix} k(s_{42\parallel} \times s_{52\parallel})^T & l_{22}^T \end{bmatrix}^T \tag{A.18} $$

$$ w_{c2} = \begin{bmatrix} 0^T & s_{12\parallel}^T \end{bmatrix}^T \tag{A.19} $$

*A.3. Inverse rate kinematics of T-mechanism*

By combining all the limb relations, the inverse rate kinematics of the T-mechanism can be obtained. The inverse Jacobian shown in Eq. (14) is obtained by multiplying the joint wrench screws Eq. (A.9), Eq. (A.10), Eq. (A.18), and Eq. (A.19) by the extended forward Jacobian Eq. (A.6) and Eq. (A.13) premultiplied by the transformation matrix Eq. (A.8), respectively, with some rearrangements.

For example, Eq. (A.9) is multiplied by the transformed extended Jacobian to obtain:

$$ w_{as}^T M J_{es} = \begin{bmatrix} d_{s1} & d_{s2} & d_{s3} & d_{s4} & d_{s5} & d_{s6} \end{bmatrix} \tag{A.20} $$

where

$$ d_{s1} = (l_{2s} \times a_s) \cdot s_{1s\parallel} + l_{2s} \cdot (s_{1s\parallel} \times (r_{1s} + a_s)) = l_{2s} \cdot (s_{1s\parallel} \times (l_{1s} + l_{2s})) = l_{2s} \cdot (s_{1s\parallel} \times l_{1s}) \tag{A.21} $$

$$ d_{s2} = (l_{2s} \times a_s) \cdot s_{2s\parallel} + l_{2s} \cdot (s_{2s\parallel} \times (l_{2s} + a_s)) = 0 \tag{A.22} $$

$$ d_{s3} = (l_{2s} \times a_s) \cdot s_{3s\parallel} + l_{2s} \cdot (s_{3s\parallel} \times a_s) = 0 \tag{A.23} $$

$$ d_{s4} = (l_{2s} \times a_s) \cdot s_{4s\parallel} + l_{2s} \cdot (s_{4s\parallel} \times a_s) = 0 \tag{A.24} $$

$$ d_{s5} = (l_{2s} \times a_s) \cdot s_{5s\parallel} + l_{2s} \cdot (s_{5s\parallel} \times a_s) = 0 \tag{A.25} $$

$$ d_{s6} = l_{2s} \cdot s_{1s\parallel} = 0 \quad \because l_{2s} \perp s_{1s\parallel} \tag{A.26} $$

Consequently, the inverse rate kinematics for the active joint of limb $s = 1, 3$ becomes:

$$ w_{as}^T \dot{x} = w_{as}^T M J_{es} \dot{q}_s = d_{s1} \dot{q}_{1s} \tag{A.27} $$

$$ \frac{w_{as}^T}{d_{s1}} \dot{x} = \dot{q}_{1s} \tag{A.28} $$

where the moving plate velocity is $\dot{x} = [w^T; v^T]^T$ and $\dot{q}_s = [\dot{q}_{1s}, \ldots, \dot{q}_{5s}]^T$ for the joint rate. Other inverse relations are similarly obtained as follows:

$$\begin{bmatrix} \dfrac{(l_{21} \times a_1)^T}{l_{21} \cdot (s_{11\parallel} \times l_{11})} & \dfrac{l_{21}^T}{l_{21} \cdot (s_{11\parallel} \times l_{11})} \\ k\dfrac{(s_{42\parallel} \times s_{52\parallel})^T}{s_{12\parallel} \cdot (l_{12} \times l_{22})} & \dfrac{l_{22}^T}{s_{12\parallel} \cdot (l_{12} \times l_{22})} \\ \dfrac{(l_{23} \times a_3)^T}{l_{23} \cdot (s_{13\parallel} \times l_{13})} & \dfrac{l_{23}^T}{l_{23} \cdot (s_{13\parallel} \times l_{13})} \\ (s_{11\parallel} \times a_1)^T & s_{11\parallel}^T \\ \mathbf{0}^T & s_{12\parallel}^T \\ (s_{13\parallel} \times a_3)^T & s_{13\parallel}^T \end{bmatrix} \begin{bmatrix} w \\ v \end{bmatrix} = \begin{bmatrix} \dot{q}_{11} \\ \dot{q}_{12} \\ \dot{q}_{13} \\ 0 \\ 0 \\ 0 \end{bmatrix} \quad (A.29)$$

where the first three rows are the active joints and the remaining rows are the constraints of each limb.

**Appendix B. Selection matrix for 1T2R PMs with nominal velocity**

As it is discussed in Section 2, establishment of selection matrix is dependent on the relation between linear velocity of points ($v_i$) and moving plate ($\dot{x}$). However, PMs different from $T_x R_y R_z, T_y R_x R_z$ and $T_z R_x R_y$, cannot be described with one component velocity. In such cases, combining velocity components from different points might be required to fully describe the moving plate's motion. Here after, we will call the combined velocity *"nominal velocity"*. Usually, selection matrices have entries of 1s and 0s. However, for a group of PMs described with nominal velocity, a selection matrix should include some geometric parameters to eliminate unwanted motions. Consequently, we refer to this selection matrix as an "*extended selection matrix.*"

For example, consider the $T_x R_x R_y$ PMs, for which the desired variables that describe their motion are $v_x$, $\omega_x$, and $\omega_y$. However, $v_i$ in Eq. (B.1) includes only some of the desired motion marked by the underbar.

$$\begin{aligned} v_{ix} &= \underline{v_x} + \omega_y a_{iz} - \omega_z a_{iy} \\ v_{iy} &= v_y - \underline{\omega_x} a_{iz} + \omega_z a_{ix} \\ v_{iz} &= v_z + \underline{\omega_x} a_{iy} - \underline{\omega_y} a_{ix} \end{aligned} \quad (B.1)$$

Therefore, in this case, we need to use nominal velocity by combining $v_{ix}$ and $v_{iz}$ components from the points. This approach allows us to incorporate all independent motion of the moving plate and remove any unwanted motions involved. The resulting nominal velocity has all information of the moving plate and can fully describe its motion.

For the $T_x R_x R_y$ PM mentioned above, we can have the following extended selection matrix ($\underline{S}$) with the condition that $v_x, \omega_x, \omega_y$ must be included while removing $v_z$, $v_y$ and $\omega_z$.

$$\underline{S} = \begin{bmatrix} \dfrac{a_{1y}}{a_{2y} - a_{1y}} & 0 & 1 & -\dfrac{a_{1y}}{a_{2y} - a_{1y}} & 0 & -1 & 0 & 0 & 0 \\ 0 & 0 & 0 & \dfrac{a_{3y}}{a_{3y} - a_{2y}} & 0 & 1 & -\dfrac{a_{2y}}{a_{3y} - a_{2y}} & 0 & -1 \\ -\dfrac{a_{3y}}{a_{1y} - a_{3y}} & 0 & -1 & 0 & 0 & 0 & \dfrac{a_{1y}}{a_{1y} - a_{3y}} & 0 & 1 \end{bmatrix} \quad (B.2)$$

Note that $\underline{S}$ has no dimension, meaning that it is equivalent to the standard selection matrix in terms of its units. By multiplying the extended selection matrix given in Eq. (B.2) with Eq. (5), we obtain a $3 \times 6$ matrix that maps the nominal velocity on the platform to the task velocity.

Thus, the equation for the nominal velocity on the moving plate becomes:

$$\begin{bmatrix} \underline{v_1} \\ \underline{v_2} \\ \underline{v_3} \end{bmatrix} = \begin{bmatrix} 1 & (a_{1y} - a_{2y}) & -\left(a_{1x} - a_{2x} - \dfrac{a_{1y}a_{2z}}{a_{1y} - a_{2y}} + \dfrac{a_{2y}a_{1z}}{a_{1y} - a_{2y}}\right) \\ 1 & (a_{2y} - a_{3y}) & -\left(a_{2x} - a_{3x} - \dfrac{a_{2y}a_{3z}}{a_{2y} - a_{3y}} + \dfrac{a_{3y}a_{2z}}{a_{2y} - a_{3y}}\right) \\ 1 & (a_{3y} - a_{1y}) & -\left(a_{3x} - a_{1x} - \dfrac{a_{1y}a_{3z}}{a_{1y} - a_{3y}} + \dfrac{a_{3y}a_{1z}}{a_{1y} - a_{3y}}\right) \end{bmatrix} \begin{bmatrix} v_x \\ \omega_x \\ \omega_y \end{bmatrix} \quad (B.3)$$

where

$$\begin{aligned} \underline{v_1} &= v_{1z} - v_{2z} + \dfrac{a_{1y}}{a_{1y} - a_{2y}} v_{2x} - \dfrac{a_{2y}}{a_{1y} - a_{2y}} v_{1x} \\ \underline{v_2} &= v_{2z} - v_{3z} + \dfrac{a_{2y}}{a_{2y} - a_{3y}} v_{3x} - \dfrac{a_{3y}}{a_{2y} - a_{3y}} v_{2x} \\ \underline{v_3} &= v_{3z} - v_{1z} + \dfrac{a_{1y}}{a_{1y} - a_{3y}} v_{3x} - \dfrac{a_{3y}}{a_{1y} - a_{3y}} v_{1x} \end{aligned}$$

**Table B.6**
Enumeration of a variety of 1T2R parallel manipulators and corresponding point motion.

| Variety of 1T2R PMs | Type of representative motion | Desired motion of MP[a] | Undesired motion of MP[a] |
|---|---|---|---|
| $T_x R_x R_y$ | Combination | $v_x, \omega_x, \omega_y$ | $v_y, v_z, \omega_z$ |
| $T_x R_x R_z$ | Combination | $v_x, \omega_x, \omega_z$ | $v_y, v_z, \omega_y$ |
| $T_x R_y R_z$ | $v_{ix}$ | $v_x, \omega_y, \omega_z$ | $v_y, v_z, \omega_x$ |
| $T_y R_x R_y$ | Combination | $v_y, \omega_x, \omega_y$ | $v_x, v_z, \omega_z$ |
| $T_y R_x R_z$ | $v_{iy}$ | $v_y, \omega_x, \omega_z$ | $v_x, v_z, \omega_y$ |
| $T_y R_y R_z$ | Combination | $v_y, \omega_y, \omega_z$ | $v_x, v_z, \omega_x$ |
| $T_z R_x R_y$ | $v_{iz}$ | $v_z, \omega_x, \omega_y$ | $v_x, v_y, \omega_z$ |
| $T_z R_x R_z$ | Combination | $v_z, \omega_x, \omega_z$ | $v_x, v_y, \omega_y$ |
| $T_z R_y R_z$ | Combination | $v_z, \omega_y, \omega_z$ | $v_x, v_y, \omega_x$ |

[a]MP stands for moving platform.

In Eq. (B.3), we can see that $v_x, \omega_x$ and $\omega_y$ are included and the undesired motion $v_z$ and $\omega_z$ are removed. Thus, the points nominal velocity is mapped to the three independent motion, $v_x, \omega_x,$ and $\omega_y,$ of the moving plate. Following the same procedure, the extended selection matrix and nominal velocity of the rest of 1T2R manipulators can be obtained. Nominal velocity is an independent velocity that fully represent the motion of the moving plate.

Table B.6 lists all the different types of 1T2R PMs, along with a single and combined component velocities used to describe the motion of the moving plate during the formulation of the dimensionally homogeneous Jacobian. The table also includes the independent parameters of each manipulator. By referring to the information provided in the table, as well as the descriptions and examples presented above, it is possible to establish the selection matrix for other 1T2R manipulators.

## References

[1] J.A. Carretero, R.P. Podhorodeski, M.A. Nahon, C.M. Gosselin, Kinematic analysis and optimization of a new three degree-of-freedom spatial parallel manipulator, Trans. ASME, J. Mech. Des. 122 (1) (2000) 17–24, http://dx.doi.org/10.1115/1.533542.
[2] Q. Li, J.M. Hervé, 1T2r parallel mechanisms without parasitic motion, IEEE Trans. Robot. 26 (3) (2010) 401–410, http://dx.doi.org/10.1109/TRO.2010.2047528.
[3] Q. Li, Z. Chen, Q. Chen, C. Wu, X. Hu, Parasitic motion comparison of 3-PRS parallel mechanism with different limb arrangements, Robot. Comput.-Integr. Manuf. 27 (2) (2011) 389–396, http://dx.doi.org/10.1016/j.rcim.2010.08.007.
[4] H. Nigatu, D. Kim, Optimization of 3-dof manipulators' parasitic motion with the instantaneous restriction space-based analytic coupling relation, Appl. Sci. (Switzerland) 11 (10) (2021) 1–22, http://dx.doi.org/10.3390/app11104690.
[5] C. Yang, W. Ye, Q. Li, Review of the performance optimization of parallel manipulators, Mech. Mach. Theory 170 (November 2021) (2022) 104725, http://dx.doi.org/10.1016/j.mechmachtheory.2022.104725.
[6] A. Rosyid, B. El-Khasawneh, A. Alazzam, Review article: Performance measures of parallel kinematics manipulators, Mech. Sci. 11 (1) (2020) 49–73, http://dx.doi.org/10.5194/ms-11-49-2020.
[7] Z. Shao, X. Tang, L. Wang, D. Sun, Atlas based kinematic optimum design of the Stewart parallel manipulator, Chin. J. Mech. Eng. 28 (1) (2015) 20–28, http://dx.doi.org/10.3901/CJME.2014.0929.155.
[8] C. Gosselin, J. Angeles, The optimum kinematic design of a spherical three-degree-of-freedom parallel manipulator, J. Mech. Transm. Autom. Des. 111 (2) (1989) 202–207, http://dx.doi.org/10.1115/1.3258984.
[9] K.H. Pittens, R.P. Podhorodeski, A family of stewart platforms with optimal dexterity, J. Robot. Syst. 10 (4) (1993) 463–479, http://dx.doi.org/10.1002/rob.4620100405.
[10] C. Reinaldo, S.N. Phu, T. Essomba, L. Nurahmi, Kinematic comparisons of hybrid mechanisms for bone surgery: 3-PRP-3-RPS and 3-RPS-3-PRP, Machines 10 (11) (2022) 1–19, http://dx.doi.org/10.3390/machines10110979.
[11] M. Tandirci, J. Angeles, F. Ranjbaran, The characteristic point and the characteristic length of robotic manipulators, in: 22nd Biennial Mechanisms Conference: Robotics, Spatial Mechanisms, and Mechanical Systems, American Society of Mechanical Engineers, 1992, pp. 203–208, http://dx.doi.org/10.1115/DETC1992-0216.
[12] L. Stocco, S. Salcudean, F. Sassani, On the use of scaling matrices for task-specific robot design, IEEE Trans. Robot. Autom. 15 (5) (1999) 958–965, http://dx.doi.org/10.1109/70.795800.
[13] W.A. Khan, J. Angeles, The kinetostatic optimization of robotic manipulators: The inverse and the direct problems, J. Mech. Des. 128 (1) (2006) 168–178, http://dx.doi.org/10.1115/1.2120808.
[14] H. Liu, T. Huang, D.G. Chetwynd, A method to formulate a dimensionally homogeneous Jacobian of parallel manipulators, IEEE Trans. Robot. 27 (1) (2011) 150–156, http://dx.doi.org/10.1109/TRO.2010.2082091.
[15] G. Pond, J.A. Carretero, Formulating Jacobian matrices for the dexterity analysis of parallel manipulators, Mech. Mach. Theory 41 (12) (2006) 1505–1519, http://dx.doi.org/10.1016/j.mechmachtheory.2006.01.003.
[16] G. Pond, J.A. Carretero, Dexterity measures and their use in quantitative dexterity comparisons, Meccanica 46 (1) (2011) 51–64, http://dx.doi.org/10.1007/s11012-010-9381-1.
[17] Z. Liu, J. Angeles, The properties of constant-branch four-bar linkages and their applications, J. Mech. Des. 114 (4) (1992) 574–579, http://dx.doi.org/10.1115/1.2917046.
[18] G. Sutherland, B. Roth, A transmission index for spatial lcclanisms, Trans. ASME, J. Manuf. Sci. Eng. 95 (2) (1973) 589–597, http://dx.doi.org/10.1115/1.3438195.
[19] J. Wang, C. Wu, X.J. Liu, Performance evaluation of parallel manipulators: Motion/force transmissibility and its index, Mech. Mach. Theory 45 (10) (2010) 1462–1476, http://dx.doi.org/10.1016/j.mechmachtheory.2010.05.001.
[20] Z.F. Shao, J. Mo, X.Q. Tang, L.P. Wang, Transmission index research of parallel manipulators based on matrix orthogonal degree, Chin. J. Mech. Eng. (Engl. Ed.) 30 (6) (2017) 1396–1405, http://dx.doi.org/10.1007/s10033-017-0193-2.

[21] J.P. Merlet, Jacobian, manipulability, condition number and accuracy of parallel robots, in: Robotics Research, Vol. 28, Springer Berlin Heidelberg, Berlin, Heidelberg, 2007, pp. 175–184, http://dx.doi.org/10.1007/978-3-540-48113-3_16.
[22] H. Liu, T. Huang, A. Kecskeméthy, D.G. Chetwynd, Q. Li, Force/motion transmissibility analyses of redundantly actuated and overconstrained parallel manipulators, Mech. Mach. Theory 109 (December 2016) (2017) 126–138, http://dx.doi.org/10.1016/j.mechmachtheory.2016.11.011.
[23] J. Angeles, Is there a characteristic length of a rigid-body displacement? Mech. Mach. Theory 41 (8) (2006) 884–896, http://dx.doi.org/10.1016/j.mechmachtheory.2006.03.010.
[24] H. Lipkin, J. Duffy, Hybrid twist and wrench control for a robotic manipulator, J. Mech. Transm. Autom. Des. 110 (2) (1988) 138–144, http://dx.doi.org/10.1115/1.3258918.
[25] K.L. Doty, C. Melchiorri, C. Bonivento, A theory of generalized inverses applied to robotics, Int. J. Robot. Res. 12 (1) (1993) 1–19, http://dx.doi.org/10.1177/027836499301200101.
[26] K.L. Doty, E.M. Schwartz, C. Melchiorri, C. Bonivento, Robot manipulability, IEEE Trans. Robot. Autom. 11 (3) (1995) 462–468, http://dx.doi.org/10.1109/70.388791.
[27] A. Fattah, A. Hasan Ghasemi, Isotropic design of spatial parallel manipulators, Int. J. Robot. Res. 21 (9) (2002) 811–824, http://dx.doi.org/10.1177/0278364902021009842.
[28] C.M. Gosselin, The optimum design of robotic manipulators using dexterity indices, Robot. Auton. Syst. 9 (4) (1992) 213–226, http://dx.doi.org/10.1016/0921-8890(92)90039-2.
[29] S.G. Kim, J. Ryu, New dimensionally homogeneous Jacobian matrix formulation by three end-effector points for optimal design of parallel manipulators, IEEE Trans. Robot. Autom. 19 (4) (2003) 731–737, http://dx.doi.org/10.1109/TRA.2003.814496.
[30] D. Kim, W.K. Chung, Analytic formulation of reciprocal screws and its application to nonredundant robot manipulators, Trans. ASME, J. Mech. Des. 125 (1) (2003) 158–164, http://dx.doi.org/10.1115/1.1539508.
[31] H. Lipkin, Material and local time derivatives of screws with applications to dynamics and stiffness, Proc. ASME Des. Eng. Tech. Conf. 2 B (1999) (2004) 1339–1348, http://dx.doi.org/10.1115/detc2004-57506.
[32] H. Nigatu, D. Kim, The T-ROBOT : a parasitic motion free 1T2R parallel manipulator, in: IT Convergence KSME Spring Conference, Andong, Korea, 2022, http://dx.doi.org/10.13140/RG.2.2.31968.15364/2.
[33] T.T. Lu, S.H. Shiou, Inverses of 2 × 2 block matrices, Comput. Math. Appl. 43 (1–2) (2002) 119–129, http://dx.doi.org/10.1016/S0898-1221(01)00278-4.
[34] I.A. Bonev, J. Ryu, New approach to orientation workspace analysis of 6-DOF parallel manipulators, Mech. Mach. Theory 36 (1) (2001) 15–28, http://dx.doi.org/10.1016/S0094-114X(00)00032-X.
[35] O. Masory, J. Wang, Workspace evaluation of Stewart platforms, in: 22nd Biennial Mechanisms Conference: Robotics, Spatial Mechanisms, and Mechanical Systems, Vol. Part F1680, American Society of Mechanical Engineers, 1992, pp. 337–346, http://dx.doi.org/10.1115/DETC1992-0232.
[36] J.A. Carretero, M.A. Nahon, R.P. Podhorodeski, Workspace analysis and optimization of a novel 3-DOF parallel manipulator, Int. J. Robot. Autom. 15 (4) (2000) 178–188.
[37] C. Yang, Q. Li, W. Ye, Dimensional synthesis method of parallel manipulators based on the principle component analysis, Mech. Mach. Theory 176 (June) (2022) 104980, http://dx.doi.org/10.1016/j.mechmachtheory.2022.104980.
[38] H. Nigatu, Y.H. Choi, D. Kim, Analysis of parasitic motion with the constraint embedded Jacobian for a 3-PRS parallel manipulator, Mech. Mach. Theory 164 (2021) 104409, http://dx.doi.org/10.1016/j.mechmachtheory.2021.104409.
[39] G. Pond, J.A. Carretero, Architecture optimisation of three 3-under(P, combining low line)RS variants for parallel kinematic machining, Robot. Comput.-Integr. Manuf. 25 (1) (2009) 64–72, http://dx.doi.org/10.1016/j.rcim.2007.09.002.
[40] D. Kim, Kinematic Analysis of Spatial Parallel Manipulators: Analytic Approach with The Restriction Space (Dissertation), POSTECH, 2002, p. 272.
[41] H. Nigatu, Y. Ho Choi, D. Kim, On the structural constraint and motion of 3-PRS parallel kinematic machines, in: Volume 8A: 45th Mechanisms and Robotics Conference (MR), Vol. 8A-2021, American Society of Mechanical Engineers, 2021, http://dx.doi.org/10.1115/DETC2021-70160.